# Evaluating Large Language Models for Anxiety and Depression Classification using Counseling and Psychotherapy Transcripts

Junwei Sun[1,+], Siqi Ma[1,+], Yiran Fan[1,+], Peter Washington[2,*]

[1]Stanford University, Stanford, CA, USA
[2]University of Hawaii at Manoa, Honolulu, HI, USA

*Correspondence should be sent to: pyw@hawaii.edu
[+]These authors contributed equally to this work.

## Abstract

We aim to evaluate the efficacy of traditional machine learning and large language models (LLMs) in classifying anxiety and depression from long conversational transcripts. We fine-tuned both established transformer models (BERT, RoBERTa, Longformer) and more recent large models (Mistral-7B), trained a Support Vector Machine with feature engineering, and assessed GPT models through prompting. We observe that state-of-the-art models fail to enhance classification outcomes compared to traditional machine learning methods.

## 1 Introduction

Identifying psychological symptoms is a crucial initial step in psychological interventions. With the rise of Large Language Models (LLMs) and their ability to facilitate conversations, researchers have been inspired to explore their potential for providing automatic interventions [1, 2, 3, 4, 5]. This focus on using machine learning (ML) methods to detect mental health issues in a scalable and accessible manner has grown alongside the increasing prevalence of mental health concerns.

There has been a recent increase in the use of natural language processing (NLP) techniques for mental illness detection, with a majority of such research employing traditional ML models like Support Vector Machines (SVM), decision trees, and random forests [6, 7, 8]. With advancements in deep learning (DL) techniques and the advent of LLMs, neural networks that rely less on feature engineering and instead can capture more syntactic nuances are also gaining importance in mental illness detection [9]. However, most studies utilize short or mixed-quality web text corpora such as tweets and Reddit posts as surrogates for mental health states, which may not fully capture the complexity of mental illness diagnostics [10, 11, 12, 13, 14]. For applications of LLMs with clinical data, most studies have utilized electronic health records (EHRs) [15, 16, 17]. However, these records may be challenging to generalize for mental health symptom detection due to the inherently complex nature of diagnosing mental health conditions. In particular, whether these popular frameworks can achieve accurate mental health diagnostics in lengthy, conversational-style psychological transcripts is not fully explored. Part of the bottleneck lies in the neural architectures' limitation in input token sizes, and recent studies have suggested methods such as text segmentation, sliding windows, and architectural adjustments like incorporating convolutional layers to address this issue [18, 19, 20].

In this study, we aim to evaluate the efficacy of applying DL techniques to classify anxiety and depression from long counseling and psychotherapy transcripts. We employ various neural network architectures and approaches, including document truncation and sub-document splitting with boosting, to build multi-label classification models capable of processing long text inputs. We ultimately achieve negative results, suggesting that current LLMs do not yet provide significantly better classification performance compared to traditional ML methods augmented with feature engineering for complex and subjective psychiatric prediction tasks using long text sequences.

## 2 Methods

**Dataset** This study utilizes data sourced from *Alexander Street Press: Counseling and Psychotherapy Transcripts, Volume I and II*, acquired from the Stanford Library [21]. The dataset comprises de-identified plain-text transcripts of therapy sessions addressing a diverse array of mental health issues with various therapeutic approaches. We applied several preprocessing steps to the transcripts: 1) non-ASCII characters were systematically eliminated; 2) descriptive elements such as "chuckles" and "laughter" were removed; and 3) explicit mentions of the symptom words (e.g., "anxiety," "depression") were removed to avoid directly associating symptom labels with classification outcomes. The preprocessed dataset consisted of 3,503 session records. Subsequently, we divided the 3,503 psychotherapy sessions into an 80% training set and a 20% evaluation set for performance assessment. An overview of how the dataset is used for downstream model training is presented in Figure 1.

**ML and Human Baselines** We established a traditional ML baseline using radial basis kernel Support Vector Machines (RBF SVMs) with a feature matrix composed of normalized stemming Bag-of-Words (BoW) and features derived from established linguistic dictionaries. These included the average concreteness score, which evaluates the degree to which the word describes a perceptible concept, and eight basic emotions and sentiments such as anger, sadness, fear, etc. calculated per sentence and averaged over each document [22, 23]. The final feature matrix consists of $30,770$ features. To assess human baseline performance, we randomly selected 100 examples from the dataset and classified whether conversation participants showed symptoms of anxiety or depression. The fine-tuned models learned psychiatry-specific knowledge within conversational contexts only, so their learning was limited to a finite set of non-expert information, such as medications and typical symptoms of various mental health states. Our annotation process mirrored this learning paradigm without strong psychiatry domain-specific knowledge to ensure comparability in performance metrics.

**Transformer Finetuning** We approached DL methods under two schemas: truncation and subdocument slicing followed by pooling. The truncation method processes the first 512 or 4096 tokens only (i.e., longest accepted sequence length) from each sample to fine-tune multi-label classifiers using BERT, RoBERTa, and Longformer models [24, 25, 26]. These models were selected due to their unique training and attention mechanisms suitable for handling varying text lengths and complexities. Additionally, we explored the efficacy of sub-document slicing and pooling by employing a boosting methodology where classification results from sliced sub-samples were pooled and subjected to either a majority vote or an OR construction (any true sub-document makes the entire document true) to determine the final predictions. Each model featured a classification head with linear, dropout, and tanh layers to produce logit outputs, facilitated by the PyTorch framework and leveraging pre-trained models from Huggingface [27, 28]. Furthermore, we also fine-tuned the more recent Mistral-7B model with qLoRA [29, 30], leveraging existing code source on this multi-label task [31]. Mistral-7B accepts longer sequence lengths and has significantly more parameters, ensuring more comprehensive coverage of transcript information and introducing more complexity. With Mistral-7B, we truncated each document up to 8192 tokens, covering the complete transcript for most samples.

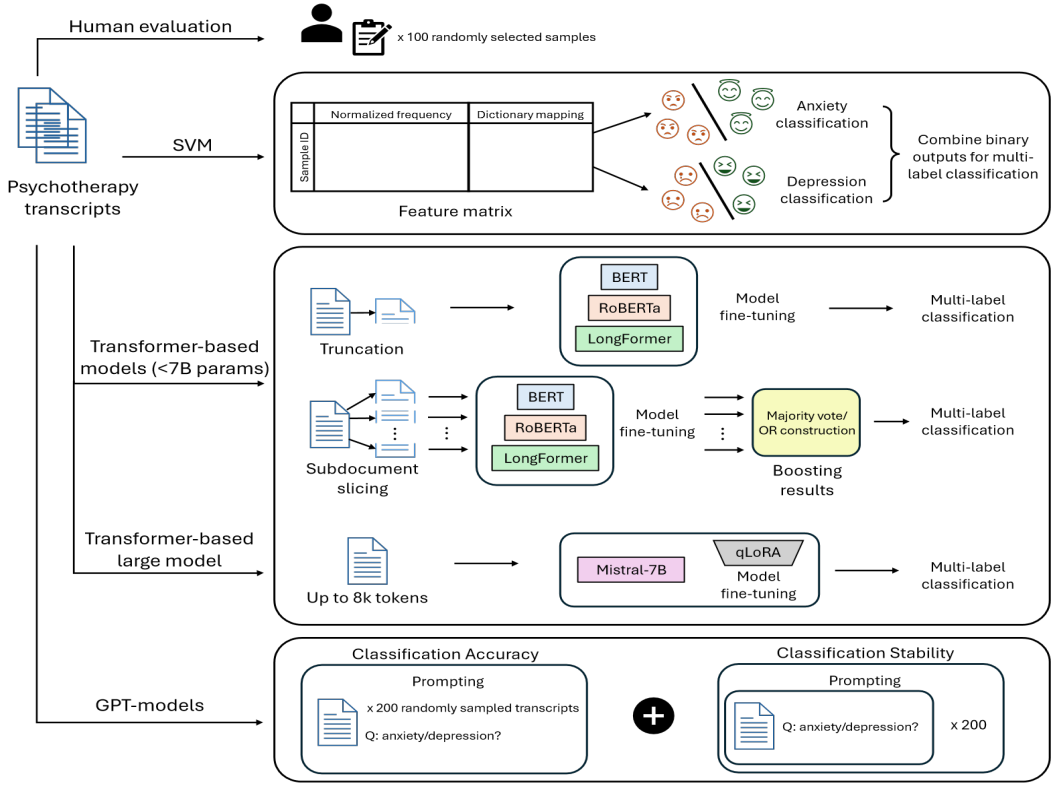

Figure 1: Overview of the methods utilized in this study. Psychotherapy transcripts are subjected to anxiety and depression classifications under: 1) Human evaluations where transcripts are manually annotated. 2) Support vector machine where models are trained on stem word frequency augmented with psychological dictionary mapping. 3) Transformer-based models fine-tuned with text truncation and subdocument splitting plus boosting. 4) GPT prompting through API calls.

**GPT Evaluations** The advent of AI systems facilitated by LLMs suggests exciting new possibilities in breaking down complex psychological texts using multi-billion scale models. We explored the current capabilities of GPT-family models, including GPT-3.5, GPT-4, and GPT-4o, in classifying psychological symptoms from text transcripts through prompting using API access. Performance is assessed from two aspects: accuracy and stability. For accuracy assessment, 200 randomly selected transcripts are subject to binary classification of anxiety and depression under a designed prompt and function calling that restricts the output to binary. Results are then pooled for multi-label accuracy. For stability assessment, a randomly sampled transcript is classified by GPT under the same prompt 200 times for each symptom and each time the multi-label classification outcome is recorded.

## 3  Results

Table 1 summarizes the performance of each baseline and DL model. Our results suggest that DL approaches, including both truncation and sub-document slicing, did not significantly outperform the traditional ML method and human baseline. The human baseline resembles DL models in terms of F1 and AUROC scores, and it is important to note that the baseline is established by non-experts. Under such circumstances, the DL systems only show an average accuracy of 0.1 higher. The performance disparities between traditional ML and DL approaches are also not substantial despite the slight improvements in F1 and AUROC. This indicates that traditional ML models with meticulous feature engineering that borrows from prior psychological research may still achieve comparable results to large neural networks in this task. This finding contradicts our hypothesis that neural networks, which convert words into word embeddings, would better capture word-level patterns and sentence-level semantics.

| Model | Learning rate | Accuracy | Weighted F1 | Weighted AUROC |
|---|---|---|---|---|
| Human Evaluation | - | 0.490 | 0.529 | 0.656 |
| RBF SVM | - | 0.549 | 0.485 | 0.705 |
| BERT | 5e-5 | 0.503 | 0.508 | 0.694 |
| Boosted BERT | 5e-5 | 0.530 | 0.351 | 0.686 |
| RoBERTa | 1e-5 | 0.561 | 0.517 | 0.713 |
| Boosted RoBERTa | 1e-5 | **0.566** | 0.542 | **0.756** |
| Longformer | 1e-5 | 0.549 | **0.565** | 0.681 |
| Boosted Longformer | 2e-5 | 0.514 | 0.319 | 0.568 |
| Mistral-7B | 5e-6 | 0.087 | 0.483 | 0.507 |

Table 1: Performance of human baseline and DL/ML models. Highest achieving scores for each performance metric are bolded.

Additionally, attention mechanisms that allow the model to associate words with their prior contexts did not seem to boost neural networks' performance. However, it is possible that the characteristics of lengthy texts render the classification task challenging for neural network-based models, as these models must infer a close-to-binary answer from a vast amount of information that may contain noise. The performance of our DL models is consistent with existing literature where a fine-tuned Clinical-Longformer tailored to long document classification tasks achieved an F1-score of 0.484 and an AUROC score of 0.762 in predicting acute kidney injury using electronic health records [32].

While fine-tuning Mistral-7B often yields desirable results within 10 epochs, utilizing Mistral on mental health tasks showed no significant learning throughout the training process. The best model also did not outperform SVM or other DL approaches. Our result is consistent with previous literature on stance classification regarding climate change activism using Mistral-7B which yielded slightly higher performance compared to the baseline model [33].

| Model | Accuracy | Weighted F1 | Weighted AUROC |
|---|---|---|---|
| Default Truncation BERT | 0.503 | 0.508 | 0.694 |
| Majority Vote | 0.483 | 0.475 | 0.647 |
| OR Construction | 0.455 | 0.439 | 0.627 |
| Random Truncation | 0.514 | 0.344 | 0.663 |

Table 2: Performance of best fine-tuned BERT-based models.

We also conducted further experiments to fine-tune the BERT model on a random segment of 512 adjacent tokens from each document and altered the pooling mechanism to OR construction. The results, illustrated in Table 2, suggest no difference compared to the truncation baseline, which may imply that information in a therapy session is uniformly scattered throughout the conversation. This aligns with previous literature on emotion classification, where limited improvement of ensemble methods on deep neural network classifiers was found [34, 35]. Specifically, simple ensemble methods with deep neural networks may help to digest longer text but fail to provide substantial performance improvement. These results from various DL experiments reinforce our main finding that DL methods do not perform as well as expected in this context.

| Model | Accuracy |
|---|---|
| GPT-4o | 0.345 |
| GPT-3.5 | 0.355 |

| GPT-4o | | Anxiety | |
|---|---|---|---|
| | | + | - |
| Depression | + | 200 | 0 |
| | - | 0 | 0 |

| GPT-3.5 | | Anxiety | |
|---|---|---|---|
| | | + | - |
| Depression | + | 0 | 83 |
| | - | 0 | 117 |

(a) Accuracy  (b) Stability of GPT-4o  (c) Stability of GPT 3.5

Table 3: Accuracy and stability of GPT models. Accuracy is evaluated on 200 randomly selected transcripts. Stability is assessed by classifying 1 randomly selected transcript 200 times. The selected transcript has both positive true labels.

Lastly, our explorations with GPTs (Table 3) validated that complex models do not necessarily yield superior performance, specifically on mental health tasks that involve discerning emotions from scattered conversations. When given 200 randomly selected transcripts for multi-label classification, both GPT-3.5 and the newest GPT-4o models' accuracy does not achieve the human baseline and is significantly lower than the performance of other ML/DL models (Table 3a). Additionally, we noticed the older GPT-3 model is highly unstable in its classification, with its prediction results split evenly across 2 out of the 4 possible combinations (Table 3c). Interestingly, we observed that such instability is resolved in GPT-4o, with all 200 predictions providing the same multi-label combination, which is indeed the true combination for that selected sample. We also observed a striking inconsistency in GPT-4 prompting (Table 3b), specifically the model is unable to provide a definitive classification outcome for many of the transcripts despite the same prompting and function-calling strategy being used. Consequently, we were not able to obtain reliable accuracy and stability measures for the GPT-4 model. Nevertheless, the low accuracy of even the newest GPT model suggests that current user-facing language models may not be capable of effectively performing psychiatric diagnostics at this level of complexity.

# 4 Discussion

## 4.1 Overview

Our findings indicate that fine-tuned neural networks do not significantly outperform traditional ML models in classifying lengthy therapeutic transcripts into mental state labels. We evaluated several methods, including training an RBF-SVM model with feature engineering incorporating prior psychological knowledge, fine-tuning popular transformer architectures and more recent LLMs, and prompting LLMs to assess their ability to extract psychological states from long, complex psychotherapy transcripts. Techniques such as boosting through a majority vote and OR construction did not significantly improve model performance, and larger models also did not outperform smaller transformer-based models in classifying mental health labels from long, unstructured conversational texts.

Our results provide multi-fold evidence for the conclusion that current popular DL frameworks do not surpass traditional ML models in discerning anxiety and depression from long conversational text corpora. This suggests that current large language models may not be well-suited for challenging tasks like distinguishing mood and subjectivity in this context. Traditional ML methods, such as SVMs augmented with dictionary mapping from prior psychological research, could still be the most reliable approach. This is particularly important as model interpretability is crucial when these models are used to assist in clinical diagnostics. SVMs, utilizing engineered features based on domain knowledge, offer better interpretability. Given the importance of model interpretability in helping clinicians understand the extracted information and the reasoning behind predictions [6], mental health professionals might prefer machine learning methods with careful feature engineering over deep learning or even large language models until these models are significantly improved.

## 4.2 Limitations

It is important to acknowledge several limitations in our study. Firstly, we evaluated the models using a single dataset of transcripts, all compiled from the same source. This could limit the generalizability of our findings to other psychotherapy transcripts or similar datasets. Although the dataset is large and of high transcription quality, it may not fully represent the diversity of demographics, symptoms, and therapeutic approaches present in real-world settings. Additionally, the rapid evolution of LLMs means our findings are based on the most recent and popular models available at the time of the study. Future research is needed to continuously analyze new models as they emerge. Practical implementation challenges, such as integrating these models into existing workflows and ensuring user acceptance, also need to be addressed.

## 4.3 Future Work

Despite our negative results with LLMs, there are several promising directions for future research. First, it is worthwhile to investigate additional DL architectures other than Transformer-based models to potentially improve performance in this task [36]. Moreover, expanding our framework to encompass a broader spectrum of mental health labels presents a compelling direction. The original dataset comprises a diverse array of over 60 mental health indicators, ranging from suicidal intent and sleep disturbances to hallucinations and mania, among others. Leveraging our existing framework, it is feasible and valuable to develop a more robust, symptom-rich multi-label classification system for psychological states. Furthermore, given the interdependencies often exhibited by psychological issues, incorporating a wider range of labels holds the potential to yield superior and robust predictive performance, potentially offsetting the fact that current DL models trained with extremely long text may suffer from instability [6] and deepening our understanding of the intricate relationships between various mental health manifestations.

Additionally, establishing a human-in-the-loop approach that combines human expertise with language models and DL could facilitate psychological diagnostics [37]. Creating a separate, expert human baseline for our dataset could help investigate the current performance cap in human annotations as current mental health [38]. This approach could bridge the gap between automated systems and human judgment, potentially leading to better diagnostic tools.

Finally, mental health issues such as anxiety and depression are often reflected in body language and state of mind, which are difficult to capture from textual inputs alone [39]. Incorporating data from different modalities, such as video recordings, into the current model could enhance mental health diagnostics by providing a more comprehensive input space, as has been demonstrated in other precision health domains [40] such as digital autism diagnostics [41, 42]. This multimodal approach could better capture the nuances of psychological states [6], potentially leading to more accurate and reliable assessments.

# 5 References

[1] Hornstein S, Scharfenberger J, Lueken U, et al. Predicting recurrent chat contact in a psychological intervention for the youth using natural language processing. *npj Digit Med* 2024;7:132. DOI: 10.1038/s41746-024-01121-9.

[2] Swaminathan A, López I, Garcia Mar RA, et al. Natural language processing system for rapid detection and intervention of mental health crisis chat messages. *npj Digit Med* 2023;6:213. DOI: 10.1038/s41746-023-00951-3.

[3] Balan R, Dobrean A, Poetar CR. Use of automated conversational agents in improving young population mental health: a scoping review. *npj Digit Med* 2024;7:75. DOI: 10.1038/s41746-024-01072-1.

[4] Schäfer SK, von Boros L, Schaubruch LM, et al. Digital interventions to promote psychological resilience: a systematic review and meta-analysis. *npj Digit Med* 2024;7:30. DOI: 10.1038/s41746-024-01017-8.

[5] Li H, Zhang R, Lee YC, et al. Systematic review and meta-analysis of AI-based conversational agents for promoting mental health and well-being. *npj Digit Med* 2023;6:236. DOI: 10.1038/s41746-023-00979-5.

[6] Zhang T, Schoene AM, Ji S, et al. Natural language processing applied to mental illness detection: a narrative review. *npj Digit Med* 2022;5:46. DOI: 10.1038/s41746-022-00589-7.

[7] Bayramli I, Castro V, Barak-Corren Y, et al. Predictive structured–unstructured interactions in EHR models: a case study of suicide prediction. *npj Digit Med* 2022;5:15. DOI: 10.1038/s41746-022-00558-0.

[8] Chancellor S, De Choudhury M. Methods in predictive techniques for mental health status on social media: a critical review. *npj Digit Med* 2020;3:43. DOI: 10.1038/s41746-020-0233-7.

[9] Abd-alrazaq A, Alhuwail D, Schneider J, et al. The performance of artificial intelligence-driven technologies in diagnosing mental disorders: an umbrella review. *npj Digit Med* 2022;5:87. DOI: 10.1038/s41746-022-00631-8.

[10] Mentalbert: Publicly available pretrained language models for mental healthcare. Ji S, Zhang T, Ansari L, et al. - Proceedings of the Language Resources and Evaluation Conference (LREC), 2022. https://doi.org/10.48550/arXiv.2110.15621.

[11] Salmi S, Mérelle S, Gilissen R, et al. Detecting changes in help seeker conversations on a suicide prevention helpline during the covid-19 pandemic: in-depth analysis using encoder representations from transformers. *BMC Public Health* 2022;22:530. DOI: 10.1186/s12889-022-12926-2.

[12] Su C, Xu Z, Pathak J, Wang F. Deep learning in mental health outcome research: a scoping review. *Transl Psychiatry* 2020;10. DOI: 10.1038/s41398-020-0780-3.

[13] Mangalik S, Eichstaedt JC, Giorgi S, et al. Robust language-based mental health assessments in time and space through social media. *npj Digit Med* 2024;7:109. DOI: 10.1038/s41746-024-01100-0.

[14] Kelley SW, Ní Mhaonaigh C, Burke L, et al. Machine learning of language use on twitter reveals weak and non-specific predictions. *npj Digit Med* 2022;5:35. DOI: 10.1038/s41746-022-00576-y.

[15] Huang J, Yang DM, Rong R, et al. A critical assessment of using ChatGPT for extracting structured data from clinical notes. *npj Digit Med* 2024;7:106. DOI: 10.1038/s41746-024-01079-8.

[16] Guevara M, Chen S, Thomas S, et al. Large language models to identify social determinants of health in electronic health records. *npj Digit Med* 2024;7:6. DOI: 10.1038/s41746-023-00970-0.

[17] Yang X, Chen A, PourNejatian N, et al. A large language model for electronic health records. *npj Digit Med* 2022;5:194. DOI: 10.1038/s41746-022-00742-2.

[18] Fiok K, Karwowski W, Gutierrez E, Davahli MR, Wilamowski M, Ahram T. Revisiting Text Guide, a Truncation Method for Long Text Classification. *Applied Sciences*. 2021; 11(18):8554. DOI: 10.3390/app11188554.

[19] Efficient classification of long documents using transformers. Park HH, Vyas Y, Shah K. - Proceedings of the 60th Annual Meeting of the Association for Computational Linguistics. https://doi.org/10.18653/v1/2022.acl-short.79.

[20] Chunk-bert: Boosted keyword extraction for long scientific literature via bert with chunking capabilities. Zheng Y, Cai R, Maimaiti M, et al. - 2023 IEEE 4th International Conference on Pattern Recognition and Machine Learning (PRML), 2023:385–92.https://doi.org/10.1109/PRML59573.2023.10348182.

[21] McNally A, et al. Counseling and psychotherapy transcripts, volumn I and II, 2014. DOI: 10.57761/c9da-zq22.

[22] Brysbaert M, Warriner AB, Kuperman V. Concreteness ratings for 40 thousand generally known English word lemmas. *Behav Res Methods* 2014;46:904–11. DOI: https://doi.org/10.3758/s13428-013-0403-5.

[23] Mohammad SM, Turney PD. Crowdsourcing a word–emotion association lexicon. *Comput Intell* 2013;29:436–65. DOI: https://doi.org/10.1111/j.1467-8640.2012.00460.x.

[24] Beltagy I, Peters ME, Cohan A. Longformer: The long-document transformer. arXiv preprint arXiv:2004.05150, 2020.

[25] Bert: Pre-training of deep bidirectional transformers for language understanding. Devlin J, Chang M-W, Lee K, Toutanova K - Proceedings of the 2019 Conference of the North American Chapter of the Association for Computational Linguistics: Human Language Technologies, 2018. https://doi.org/10.48550/arXiv.1810.04805.

[26] Liu Y, Ott M, Goyal N, et al. Roberta: A robustly optimized bert pretraining approach. arXiv preprint arXiv:1907.11692, 2019.

[27] Pytorch: An imperative style, high-performance deep learning library. Paszke A, Gross S, Massa F, et al. - Advances in Neural Information Processing Systems 32, NeurIPS 2019. https://doi.org/10.48550/arXiv.1912.01703.

[28] Transformers: State-of-the-art natural language processing. Wolf T, Debut L, Sanh V, et al. - Proceedings of the 2020 Conference on Empirical Methods in Natural Language Processing: System Demonstrations, 2020. https://doi.org/10.18653/v1/2020.emnlp-demos.6.

[29] QLORA: Efficient finetuning of quantized LLMs, 2023. Dettmers T, Pagnoni A, Holtzman A, Zettlemoyer L - NIPS '23: Proceedings of the 37th International Conference on Neural Information Processing Systems, 2023. https://doi.org/10.48550/arXiv.2305.14314.

[30] Jiang AQ, Sablayrolles A, Mensch A, et al. Mistral 7b, 2023. arXiv preprint arXiv: 2310.06825, 2023.

[31] Khalil N. Brev.dev. https://github.com/brevdev/notebooks/blob/main/mistral-finetune.ipynb, 2024.

[32] Li Y, Wehbe RM, Ahmad FS, Wang H, Luo Y. A comparative study of pretrained language models for long clinical text. *J Am Med Inform Assoc* 2022;30:340–7. DOI: 10.1093/jamia/ocac225.

[33] NLPDame at ClimateActivism 2024: Mistral sequence classification with PEFT for hate speech, targets and stance event detection. Christodoulou C. - Proceedings of the 7th Workshop on Challenges and Applications of Automated Extraction of Socio-political Events from Text (CASE), 2024. St. https://doi.org/10.48448/fx3t-4955

[34] Kamran S, Zall R, Hosseini S, Kangavari M, Rahmani S, Hua W. EmoDNN: understanding emotions from short texts through a deep neural network ensemble. *Neural Comput Appl* 2023;35. DOI: 10.1007/s00521-023-08435-x.

[35] Parvin T, Sharif O, Hoque MM. Multi-class textual emotion categorization using ensemble of convolutional and recurrent neural network. *SN Comput Sci* 2021;3:62. DOI: 10.1007/s42979-021-00913-0.

[36] Raiaan MAK, Mukta MSH, Fatema K, et al. A review on large language models: Architectures, applications, taxonomies, open issues and challenges. *IEEE Access* 2024. DOI: 10.1109/ACCESS.2024.3365742.

[37] Washington P. A perspective on crowdsourcing and human-in-the-loop workflows in precision health. *J Med Internet Res* 2024;26. DOI: 10.2196/51138.

[38] Van Veen D, Van Uden C, Blankemeier L, et al. Adapted large language models can outperform medical experts in clinical text summarization. *Nat Med* 2024:1–9. DOI: 10.1038/s41591-024-02855-5.

[39] Moura I, Teles A, Viana D, Marques J, Coutinho L, Silva F. Digital phenotyping of mental health using multimodal sensing of multiple situations of interest: A systematic literature review. *J Biomed Inform* 2023;138:104278. DOI: 10.1016/j.jbi.2022.104278.

[40] Kline A, Wang H, Li Y, et al. Multimodal machine learning in precision health: A scoping review. *npj Digit Med* 2022;5:171. DOI: 10.1038/s41746-022-00712-8.

[41] Washington P, Wall DP. A review of and roadmap for data science and machine learning for the neuropsychiatric phenotype of autism. *Annu Rev Biomed Data Sci* 2023;6:211–28. DOI: 10.1146/annurev-biodatasci-020722-125454.

[42] Perochon S, Di Martino JM, Carpenter KLH, et al. Early detection of autism using digital behavioral phenotyping. *Nat Med* 2023;29:2489–97. DOI: 10.1038/s41591-023-02574-3.

Address and affiliations:
JS, SM, YF: 390 Jane Stanford Way, Department of Statistics, Stanford University, Stanford, CA 94305
PW: 1680 East-West Road, Department of Information and Computer Sciences, University of Hawaii, Honolulu, HI 96822

Acknowledgment:
We would like to thank Stanford Libraries for providing access to the dataset used in this study.

Data Availability:
The data analyzed in this study are not publicly available due to access permissions from Stanford Libraries, and we are not authorized to reshare this data. However, they are available at https://redivis.com/datasets/4ew0-9qer43ndg and https://redivis.com/datasets/9tbt-5m36b443f upon reasonable request and with permission of Stanford Libraries.

Code Availability:
The repository containing the code used during the current study can be found at the following github link: https://github.com/ivysun14/Mental-Health-Prediction.